\documentclass[a4paper,conference]{IEEEtran}
\IEEEoverridecommandlockouts
\usepackage{times}
\usepackage{epsfig}
\usepackage{graphicx}
\usepackage{amsmath}
\usepackage{amssymb}
\usepackage{mathtools}
\usepackage{algorithm}
\usepackage{algorithmic}

\newcommand{\mat}[1]{\boldsymbol{#1}}
\renewcommand{\vec}[1]{\boldsymbol{#1}}

\usepackage{booktabs} % nice tables
\usepackage{color} %  mainly for commenting (see below)
\newcommand\newnotecommand[3]{%
\newcommand#1[1]{{\color{#3}\footnote{{\color{#3}#2:} ##1}}}}
\newnotecommand\joni{Joni}{red}
\newnotecommand\ugur{Ugur}{green}
\usepackage{paralist} % compactitem

\newcommand{\etal}{\textit{et al}. }

\newcommand{\eg}{\textit{e}.\textit{g}. }

\definecolor{forestgreen(web)}{rgb}{0.13, 0.55, 0.13}
\ifCLASSINFOpdf
\else
\fi
\begin{document}
\title{Depth Masked Discriminative Correlation Filter}

\author{\IEEEauthorblockN{U\u{g}ur Kart$^\star$, Joni-Kristian K\"am\"ar\"ainen$^\star$ 
\thanks{$^\star$U\u{g}ur Kart was supported by two projects: Business Finland Project "360 Video Intelligence - For Research Benefit"
with Nokia, Lynx, JJ-Net, BigHill, Leonidas
and Business Finland - FiDiPro Project "Pocket - Sized Big Visual Data"}}
\IEEEauthorblockA{$^\star$Laboratory of Signal Processing\\Tampere University of Technology\\
Email: \texttt{first.surname@tut.fi}}
\and
\IEEEauthorblockN{Ji\v{r}\'{i} Matas$^\dagger$ \thanks{$^\dagger$Ji\v{r}\'{i} Matas was supported by Czech Science Foundation
Project GACR P103/12/G084. $\copyright$ 2018 IEEE}}
\IEEEauthorblockA{$^\dagger$Faculty of Electrical Engineering\\Czech Technical University in Prague\\
Email: matas@cmp.felk.cvut.cz}
\and
\IEEEauthorblockN{Lixin Fan$^\ddagger$, Francesco Cricri$^\ddagger$}
\IEEEauthorblockA{$^\ddagger$Nokia Technologies\\Tampere, Finland\\
Email: \texttt{first.surname@nokia.com}}
}


%Telephone: (800) 555--1212, Fax: (888) 555--1212}


\maketitle

\begin{abstract}
  Depth information provides a strong cue for occlusion detection and
  handling, but has been largely omitted in generic object tracking until recently
  due to lack of suitable benchmark datasets and applications.
  In this work, we propose a {\em Depth Masked Discriminative Correlation
    Filter} (DM-DCF) which adopts novel depth segmentation based
  occlusion detection that stops correlation filter updating and
  depth masking which adaptively adjusts the spatial support for correlation
  filter. In Princeton RGBD Tracking Benchmark, our DM-DCF is among
  the state-of-the-art in overall ranking and the winner on multiple
  categories. Moreover, since it is based on DCF, ``DM-DCF'' runs an order of magnitude
  faster than its competitors making it
  suitable for time constrained applications.
\end{abstract}


% \fi
\IEEEpeerreviewmaketitle

%%%%%%%%%%%%%%%%%%%%%%%%%%%%%%%%%%%%%%%%%%%%%%%%%%%%%%%%%%%%%%%%%%%%%%%
\section{Introduction}
Generic short-term object tracking has been a popular research topic
in computer vision for the last few years and trackers based on the
{\em Discriminative Correlation Filter (DCF)}~\cite{Hester1980} have been particularly
successful for applications with time constraint~\cite{VOT2017,Henriques2014,DanelljanCVPR2017,csr}.
However, in RGB tracking, there are fundamental difficulties that can be solved with the help of depth (D) information, occlusion handling being the most obvious. 
Additionally, RGBD sensors are popular in robotics where 3D object tracking also
has many important applications, e.g., object manipulation and grasping.

\begin{figure}[h]
  \begin{center}
    \includegraphics[width=1.0\linewidth]{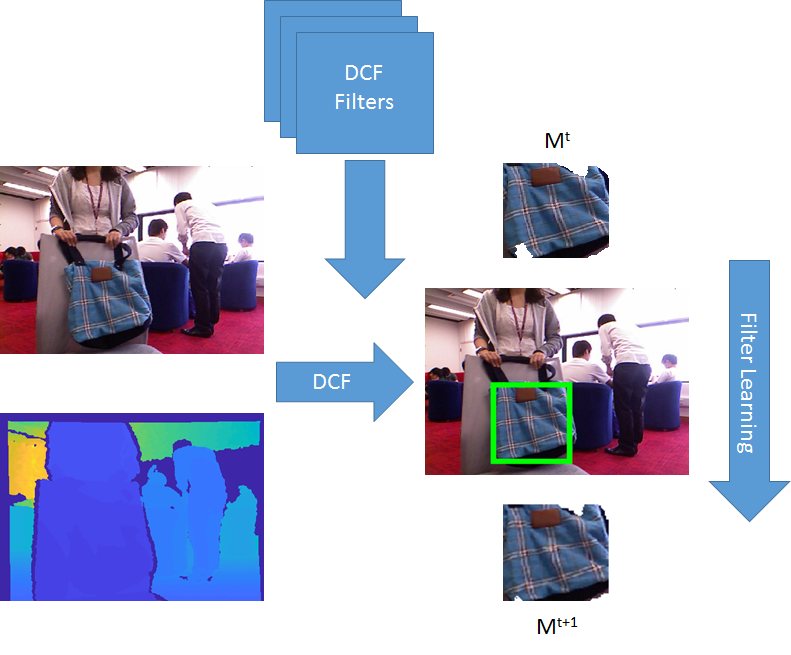}
    \caption{ Overview of our DM-DCF method. Depth cue is used
      in parallel with DCF tracking by inherently adopting depth based masks into DCF formulation
      which allows altering the filter supports. Moreover, combination of DCF and depth segmentation
      mitigates the possible errors stemming from individual observations. ($M^t$ and $M^{t+1}$ are the masks
      used for filter training where the slight differences between the masks of two consecutive frames are visible)
      \label{fig:teaser}}
  \end{center}
\end{figure}

\begin{figure}[h]
  \begin{center}
    \includegraphics[width=1.0\linewidth, height=0.79\linewidth]{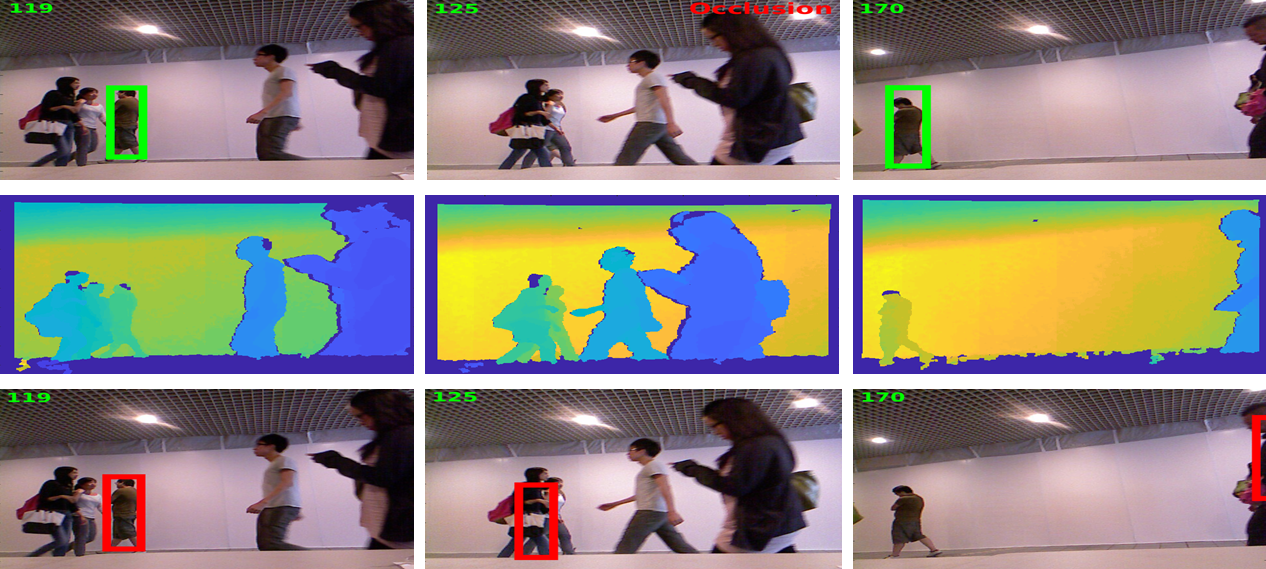}
    \caption{Example results of our DM-DCF tracker (top row) that uses depth cue (middle row)
      to construct depth masks. Depth helps detecting  occlusions as the third dimension provides a strong clue and hence, the tracker stops updating the object model. However,  RGB tracker (CSR-DCF~\cite{csr}, bottom row) is not able to detect the occlusions and continues learning occluding object and consequently, loses the target object.
      \label{fig:longtermOcc}}
  \end{center}
\end{figure}

There have been surprisingly small number of works on RGBD tracking since the
introduction of the first dedicated benchmark for RGBD tracking: Princeton
RGBD Tracking dataset~\cite{princetonrgbd}. The dataset authors also
proposed baseline trackers for 2D (depth as an additional cue) and
3D tracking (3D bounding box). More recently, particle filter based methods
have been proposed by
Meshgi~\etal~\cite{oapf} and Bibi~\etal~\cite{Bibi3D}, but they are
both slow for real-time applications. Instead, we adopt the Discriminative
Correlation Filter (DCF) approach since it is proven to be fast and successful in RGB tracking. The two other
DCF based RGBD works to our best knowledge are Hannuna et al.~\cite{DSKCFShape}
and An et al.~\cite{tlds} which are both among the top performers in the
Princeton dataset.

In aforementioned RGBD tracking methods, depth has been used as
a mere additional cue for tracking, but intuitively the most important
role for the depth information is in accurate and robust occlusion handling.
In our work, instead of separating occlusion handling and tracking, we unite the two processes
by integrating occlusion handling with correlation filters in the
sense that the spatial filter supports are dynamically altered using the
depth based segmentation.
In this work we make the following novel contributions:
\begin{compactitem}
    \item We propose a novel occlusion handling mechanism based on object's depth distribution and DCF's response history. By detecting strong occlusions, our tracker avoids corrupting the object model.
    \item We propose depth masking for DCF which avoids using the occluded regions in matching and therefore provides more reliable tracking scores.
  \item We experimentally validate Depth Masked Discriminative Correlation
    Filter (DM-DCF) tracker on
    the Princeton RGBD Tracking benchmark where it ranks among the state-of-the-art
     algorithms with ranking first in multiple categories while being faster than the other top performing  methods in the benchmark.
\end{compactitem}
Our code will be made publicly available to facilitate fair comparisons.

%%%%%%%%%%%%%%%%%%%%%%%%%%%%%%%%%%%%%%%%%%%%%%%%%%%%%%%%%%%%%%%%%%%%%%%
\section{Related Work}
\textbf{Object Tracking -- } Existing trackers for
generic short-term object tracking on RGB videos can be grouped
under two main categories, \textit{Generative} (matching with
updated target model) or \textit{Discriminative} (classifier based)
methods, depending on how a tracked object (target region) is modelled
and localized~\cite{Smeulders2014}. A generative tracker represents
the target as a collection of features from previously
tracked frames and matches them to search region in the current
frame. A few prominent examples of this family of trackers are
Incremental Visual Tracking (IVT)~\cite{IVT}, Structural Sparse
Tracking (SST)~\cite{SST} and kernel-based object tracking 
(mean shift)~\cite{meanshift}. Generative trackers build a model
from the positive examples, \eg tracked regions, but false matches
occur if the background or other objects have similar texture
properties. This issue is addressed in discriminative trackers
by continuously learning and updating a classifier that distinguishes
the target from background. In recent years, discriminative
approach has been more popular and it has produced many well-performing
trackers such as Tracking-Learning-Detection (TLD)~\cite{Kalal2011},
Continuous Convolutional Operators Tracking
(C-COT)~\cite{Danelljan2016}, Multi-Domain Convolutional Neural
Network (MDNet)~\cite{Nam2016}, ECO~\cite{DanelljanCVPR2017}, CSR-DCF~\cite{csr}. 
Due to its mathematical simplicity, efficiency and superior performance, 
we adopted the DCF based approach as our baseline to allow us to achieve fast throughput rate while 
retaining an accuracy comparable to more complicated algorithms as it is proven in 
VOT 2017~\cite{VOT2017}.

\textbf{Object Tracking with Depth -- }
The number of research papers on generic RGBD (color + depth) object
tracking is surprisingly limited despite the fact that depth sensors
are ubiquitous and the apparent application of the problem in robotics.
One of the reasons is the lack of suitable datasets until recent years. In 2013
Song \etal~\cite{princetonrgbd} introduced the Princeton Dataset for
RGBD tracking and nine variations of a baseline tracker. They
presented two main approaches; {\em depth as an additional cue}
and {\em point cloud tracking}. In the first case, depth is added
as an additional dimension to Histogram of Oriented Gradients
(HOG) feature space and in the second case tracking is based
on 3D point clouds producing also a 3D bounding box.
%outputs 3D bounding boxes. The occlusion detection on the other hand
%uses a depth histogram and assumes that the tracked object is the
%closest object in the bounding box. This assumption may prove to be
%problematic especially when the target object is partially
%occluded. On the other hand, their occlusion recovery depends on
%tracking the occluder which again assumes that the target is behind
%the same object until it reappears. In crowded scenes, it is very
%likely that the target object might move behind another
%occluder.
Their best method performs well, but contains heuristic processing stages and
its speed (0.26 FPS) makes the algorithm unsuitable
for real-time applications. 

Another RGBD method was recently proposed by Meshgi
\etal~\cite{oapf} who tackled the problem with an
occlusion-aware particle filter framework employing a probabilistic
model. Although their algorithm is among the top performers on
Princeton Benchmark, the complexity of their model, the number of
parameters to be tuned and the slowness of the algorithm (0.9 FPS)
makes it unpractical for many applications.

Bibi \etal~\cite{Bibi3D} suggested a part-based sparse tracker within
a particle filter framework. They represented particles as 3D cuboids
that consist of sparse linear combination of dictionary elements which
are not updated over the time. In case of no occlusion, their method
first finds a rough estimate of the target location using 2D optical
flow. Following this, particles are sampled in rotation $\mat{R}$
and translation $\vec{T}$ space. To detect the occlusions, they
adopted a heuristic which states that the number of points in the 3D
cuboid of the target should decrease below a threshold. Their method sets
currently the state-of-the-art on Princeton benchmark however, computation times are not mentioned in the original work.

To the authors' best knowledge there are only two RGBD trackers based on DCF
 which is used in our method.
%To the best of our knowledge, there are two algorithms in the
%literature that are similar to ours in spirit and they both are based
%on well-known Kernelized Correlation
%Filters(KCF)~\cite{Henriques2014}.
The first one was proposed by Hannuna~\etal~\cite{DSKCFShape} who use depth
for occlusion handling, scale and shape analysis. To this end, they first
apply a clustering scheme on depth histogram which is followed by
formation of a single Gaussian distribution based depth modelling where they
assume the cluster with the smallest mean must be the object (similar
heuristic used in Song~\etal~\cite{princetonrgbd}).
%In our formulation
%however, we do not have such assumptions since we let our depth model
%to do the segmentation with the help of the tracker. Thus, instead of
%assuming the object is the closest to the camera, we use the
%information coming from tracker along with the automatic depth
%segmentation.
Another shortcoming of their algorithm is that their
occlusion handling allows tracking of occluding region which
introduces same problems as in~\cite{princetonrgbd}.

Second method was proposed by An~\etal~\cite{tlds}. They
used depth based segmentation in parallel to Kernelized
Correlation Filter (KCF)~\cite{Henriques2014} detection and then interpreted the
results to locate the target and determine the occlusion state with a heuristic approach.
%On the other hand, we not only use the
%depth based segmentation as a part of occlusion detection, we also use
%it for providing better sampling masks for our tracker. Their reported
%processing time ~4.5 FPS puts them into the sub-real-time category. 

The rest of the paper is organized as follows; Section~\ref{sec:generalAlgorithm} provides an overview for the proposed
tracking method, Section~\ref{sec:experiments} reports the results of our tracker, its comparison against the state-of-the-art
RGBD trackers and also discusses the ablation studies to evaluate the impacts of our design choices. Finally Section~\ref{sec:conclusion} 
sums up our proposed method with remarks for the future work. 
The overview of the proposed method can be seen in Fig.~\ref{fig:teaser} and Alg.~\ref{algo:1}.

%%%%%%%%%%%%%%%%%%%%%%%%%%%%%%%%%%%%%%%%%%%%%%%%%%%%%%%%%%%%%%%%%%%
\section{Depth Masked DCF}\label{sec:generalAlgorithm}
Our approach is inspired by the recent work of Lukezic et
al.~\cite{csr} who robustified standard DCF by introducing filter masks and won
VOT 2017 challenge in real-time track.
However, their method is plain RGB and our RGBD tracker differs from
their work significantly in the following terms. Firstly, their method
is an RGB tracker without occlusion handling mechanism. Secondly, we
update correlation filter mask using depth cues instead of spatial
2D priors and color segmentation. As we show in Section~\ref{sec:design},
our approach is clearly superior which can be attributed to the power
of depth cue.
 
An overview of our DM-DCF algorithm is given below while
Section~\ref{sec:dcf} provides an introduction to DCF based tracking,
Section~\ref{sec:mask} reports how the depth based DCF
masks are created, Section~\ref{sec:optimization} discusses the optimization of
correlation filters with spatial constraints and Section~\ref{sec:occlusionHandling} introduces
our occlusion handling mechanism.
 
\begin{algorithm}
\caption{Depth Masked DCF}
\begin{algorithmic}
  \REQUIRE Current frame $I^{t}$;
  Occlusion state $S^{t-1}$;
  Foreground and background depth distributions $P_{fg}^{t-1}$, $P_{bg}^{t-1}$; 
  Tracker response threshold $\tau$;
  {$K$ last responses $\vec{r}$}
\IF[** Tracker part **]{$S^{t-1}$ is $false$}
\STATE Run DCF tracker ($\bf{h}^{t-1}$) on $I^{t}$
\STATE Calculate maximum filter response $r_{max}^t$
\STATE Run occlusion detection to obtain $S^{t}$
\STATE Calculate depth mask $\bf{M^t}$ (Sec.~\ref{sec:mask})
\ELSE[** Detector part **]
\STATE Run full frame detection and obtain $r_{max}^t$
\IF{$r_{max}^t > \tau * mean(\vec{r})$}
\STATE $S^{t} \leftarrow false$ 
\ELSE
\STATE $S^{t} \leftarrow true$ 
\ENDIF
\ENDIF
\IF[** Mask update part **]{$S^{t}$ is $false$}
\STATE Update distributions $P_{fg}^t$ and $P_{bg}^t$ using $\bf{M^t}$
\STATE {Update $\bf{h}^t$ using $\bf{M^t}$}
\STATE {Update tracker response history $\vec{r}^{mod(t,K)} \leftarrow r_{max}^t$}
\ENDIF
\STATE {Proceed to the next frame}
\end{algorithmic}
\label{algo:1}
\end{algorithm}

\subsection{Correlation Filters}\label{sec:dcf}
The problem to be solved in correlation filter based tracking is
to find a suitable filter $h$ at discrete time point $t$ and sample point
$i$ that provides desired output $y_i$ 
for given input image $x_i$ which includes the target object. 
Desired output, $y_i$, can be
constructed by using a small, dense, 2D gaussian at the centre of
a tracked object image~\cite{Bolme-2010-cvpr}.
Optimization of the filter can be formulated as a ridge regression problem:
\begin{equation}
  \frac{1}{2}\sum_i ||y_i-h^Tx_i||^2+\frac{\lambda}{2}||h||^2
  \label{eq:ridge}
\end{equation}
where $\lambda$ is the regularization parameter that is used to avoid
overfitting to the current object appearance. A widely used
technique is to assume circular repetitions of each input $x_i$
as $x_i(\Delta\tau_j)$ where $\Delta\tau_j$ represents
all circular shifts of $x_i$. This assumption leads to a fast filter
optimization in the Fourier domain~\cite{Henriques2014}:
\begin{equation}
  \hat{h} = \frac{\sum_i\hat{x_i}^*\odot \hat{y_i}}{\sum\hat{x_i}^*\odot \hat{x_i}+\lambda}
\end{equation}
where $\hat{h}$, $\hat{x}$ and $\hat{y}$ are the Fourier transforms
of the correlation filter, input image and the desired output respectively.
 $\odot$ is the element-wise Hadamar product and
$^*$ denotes the Hermitian conjugate. Computation in Fourier domain
reduces the complexity from $\mathcal{O}(D^3+ND^2)$ in the spatial domain
to $\mathcal{O}(ND\log D)$ for the images of the size $D$ pixels and $N$ examples as it is reported in~\cite{Galoogahi-2015-cvpr}.
However, this also enforces a special form of (\ref{eq:ridge}):
\begin{equation}
  \frac{1}{2}\sum_i \sum_j ||y_i(j)-h^Tx_i(\Delta\tau_j)||^2+\frac{\lambda}{2}||h||^2
  \label{eq:ridge2}
\end{equation}
where $j$ runs over the all $D$ circular shifts of each input $x_i$.
Henriques \etal~\cite{Henriques2014} also extended the above to
include kernel functions that can make tracker even more effective
without any loss in computation speed.

Another important part of correlation filter based tracking is time
averaging for online adaptation where previous appearances are retained in
``tracker memory''~\cite{Bolme-2010-cvpr}
\begin{equation}
\hat{h}^t = \psi\hat{h}^t + (1-\psi)\hat{h}^{t-1}
%  \begin{split}
%    \hat{h} &= \frac{\sum_i(\hat{h}_{xy})^i}{\sum_i(\hat{h}_{xx})^i+\lambda}\\
%    &(\hat{h}_{xy})^i = \eta(\hat{x}_i^* \odot \hat{y}_i)+(1-\eta)(\hat{h}_{xy})^{i-1}\\     
%    &(\hat{h}_{xx})^i = \eta(\hat{x}_i^* \odot \hat{x}_i)+(1-\eta)(\hat{h}_{xx})^{i-1}
%    \end{split} \enspace
\end{equation}

\subsection{Depth Masking}\label{sec:mask}
\begin{figure}[h]
  \begin{center}
    \includegraphics[width=1.0\linewidth]{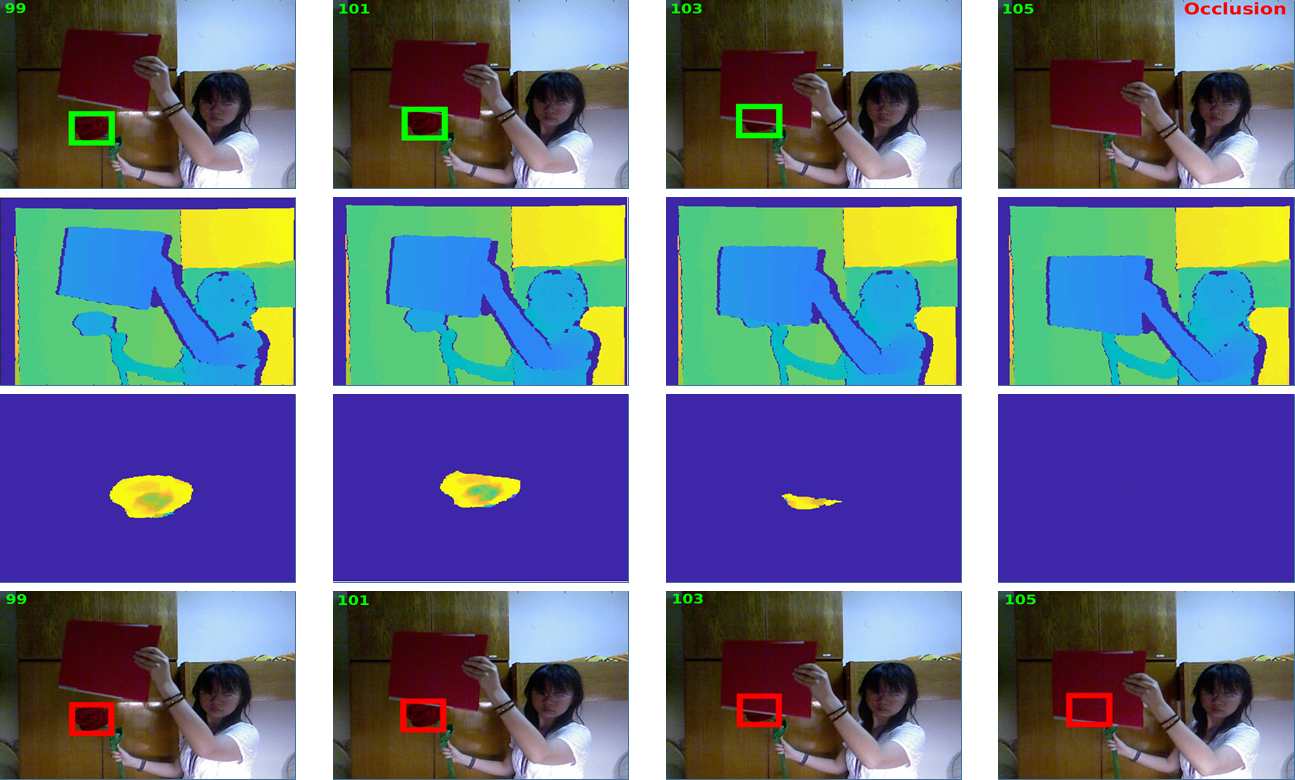}
    \caption{Our method (top) can handle the challenging cases where the state-of-the-art DCF based
     RGB tracker (CSR-DCF~\cite{csr}) fails due to the color similarity of a tracked object and back/foreground (bottom). Depth cue (second row)
     is used to generate a filter mask (third row) which is both used for filter generation and 
     occlusion handling (note detected occlusion in the last frame when detection mode is switched on).
      \label{fig:rose}}
  \end{center}
\end{figure}

The masking approach in our work to select active pixels (i.e. pixels that are used for DCF updates) for
the DCF tracker was inspired by the work of
Lukezic~\etal~\cite{csr} who constructed an RGB
cue driven mask by forming a pixel graph where the foreground was
segmented by graph cut approach using color histograms and spatial relationships.
However, their method is deemed to fail in the cases
where background and foreground are of similar color and it cannot
detect occlusions as it can be observed in Fig.~\ref{fig:longtermOcc} and Fig.~\ref{fig:rose}. On the other hand, 
in our method the depth cue
turns out to be clearly better in mask generation and also provides
an intuitive way to detect occlusion and switching from the tracking to
detection mode which provides superior performance in long-term occlusions as in Fig.~\ref{fig:longtermOcc}.
In the case of foreground masking of a tracked object, the correlation
filter is changed to a masked correlation filter, $h_M = M \odot h$, which replaces
$h$ in (\ref{eq:ridge}) or (\ref{eq:ridge2}).
The mask $M$ has value $1$ to mark
the visible (active) region of a tracked object and value $0$ to inactivate
pixels in the background. Another advantage of masked correlation is the
fact that the border effects in cyclic correlation can be removed if the
mask is made larger than the current bounding box~\cite{Galoogahi-2015-cvpr}
(up to the size of the whole input frame).

We construct our mask using probabilistic representations $P_{fg}$  and $P_{bg}$  of foreground object and background (note that background in our case
means scene elements both closer and further away from the object) respectively.
In its simplest form, the mask can be generated from foreground probability ratios
\begin{equation}
  M(\vec{x}) = \begin{cases}
    1, \hbox{if } \frac{P_{fg}}{P_{bg}} > \Omega\\
    0, \hbox{if } \frac{P_{fg}}{P_{bg}} \leq \Omega\\
    \end{cases} \enspace
\end{equation}
However, we found another approach based on adaptive thresholding to
work better since it avoids setting the threshold $\tau$. We assign
each mask pixel a probability ratio value 
$\log{\frac{P_{fg}}{P_{bg}}}$ which produces a ``foreground probablity
image'' and the probability image is thresholded to form a binary foreground
mask by the adaptive method of Otsu~\cite{Otsu-1979}.

For the probability distribution estimation we tested both single
Gaussian and Gaussian Mixture Models, but found single Gaussians
to perform better and another additional benefit is their fast online
update rules.
Our distributions are $P_{fg} \propto \mathcal{N}(\mu_{fg},\sigma^2_{fg})$
and $P_{bg} \propto \mathcal{N}(\mu_{bg},\sigma^2_{bg})$ whose
parameters are updated by the following rules:
\begin{equation}
  \begin{split}
  \mu^{(t)} = \mu^{(t)} \theta + (\mu^{(t-1)} (1-\theta)) \\
  \sigma^{(t)} = \sigma^{(t)} \gamma + (\sigma^{(t-1)} (1-\gamma))
  \end{split}
\end{equation}
where $\theta$ and $\gamma$ are fixed update rates.

To construct the new distribution for the foreground, the depth values that are in the current mask are picked. 
In the first frame however, the ground truth bounding box provided by the dataset is used to create the initial distributions.

\subsection{Filter Optimization}\label{sec:optimization}
A mask generated by the procedure in Section~\ref{sec:mask}
changes the target function (\ref{eq:ridge}) to find the optimal correlation filter $h$ into
\begin{equation}
  \frac{1}{2}\sum_i ||y_i-h^TMx_i||^2+\frac{\lambda}{2}||h||^2
  \label{eq:ridge_masked}
\end{equation}
and the circular function (\ref{eq:ridge2}) into
\begin{equation}
  \frac{1}{2}\sum_i \sum_j ||y_i(j)-Mh^Tx_i(\Delta\tau_j)||^2+\frac{\lambda}{2}||h||^2
  \label{eq:ridge2_masked}
\end{equation}
that can be written in the Fourier domain as~\cite{Galoogahi-2015-cvpr}
\begin{equation}
E(h) =  \frac{1}{2}\sum_i || \hat{y}_i - \hbox{diag}(\hat{x}_i)^T\sqrt{D}\mat{F}M^Th||^2+\frac{\lambda}{2}||h||^2
\label{eq:target}
\end{equation}
where $M^Th$ are defined in the spatial domain and $D\mat{F}$ is the Fourier
operator matrix $\mat{F}$ multiplied by the number of dimensions $D$ in the signal.
This solution is very inefficient in the Fourier
domain ($\mathcal{O}(D^3+ND^2)$) and therefore the primal solution in the spatial
is faster. However, (\ref{eq:target}) is in the form where the Lagrangian multiplier
$\hat{g}$ can be introduced and Alternating Direction Method of Multipliers (ADMM)
adopted~\cite{Boyd-2010-ADMM}:
\begin{equation}
  \begin{split}
    &\hbox{minimize } \frac{1}{2}\sum_i || \hat{y}_i - \hbox{diag}(\hat{x}_i)^T\hat{g}||^2+\frac{\lambda}{2}||h||^2\\
    &\hbox{subject to } \hat{g} =  \sqrt{D}\mat{F}M^Th
  \end{split} \enspace 
\end{equation}
The augmented Lagrangian method uses the following unconstrained objective
\begin{equation}
  \begin{split}
    &\hbox{minimize } \frac{1}{2}\sum_i || \hat{y}_i - \hbox{diag}(\hat{x}_i)^T\hat{g}||^2+\frac{\lambda}{2}||h||^2\\
    &+ \frac{\mu}{2}\left(\hat{g}-\sqrt{D}\mat{F}M^Th\right)^2
    + \xi \left(\hat{g}-\sqrt{D}\mat{F}M^Th\right)
  \end{split}
\end{equation}
where $\mu$ is the penalty term affecting the convergence and $\xi$ is the
Lagrange multiplier updated on each iteration. Optimization iteratively
updates the estimates $\hat{g}^{t+1}$ and $h^{t+1}$ and the multiplier using the rule
\begin{equation}
  \xi^{t+1} = \xi^t-\mu\left(\hat{g}^{t+1}-\sqrt{D}\mat{F}M^Th^{t+1}\right) \enspace 
\end{equation}

\subsection{Occlusion Detection}\label{sec:occlusionHandling}
Detecting heavy occlusions ($\ge 90\%$) and consequently stopping model updates is a vital
part of occlusion-aware tracking. This process allows DCF to avoid possible
model pollution which eventually leads to drifting.
In RGB-based DCF, the main source of information for occlusion handling
is a correlation filter response at the maximum location $r_{max}$ since
a rapid decrease can be considered as an evidence of
occlusion~\cite{tlds,DSKCFShape}. To include tracker based occlusion detection,
we calculate running mean of the maximum responses where the maximum
of the current frame $r_{max}^{curr}$ is added iteratively:

\begin{equation}
  r_{max}^{(t+1)} =
  r_{max}^{(t)}+\frac{r_{max}^{curr}-r_{max}^{(t)}}{t} \enspace
\end{equation}

The main drawback of the above tracker response based occlusion
detection is the implicit assumption that occlusions occur
faster than model appearance changes. This does not have to be true
and might cause false occlusion detections.
%(Figure~\ref{fig:occlusion}).
%
%\begin{figure*}
%  \begin{center}
%    \includegraphics[width=1.0\linewidth, height=0.45\linewidth]{resources/occlusion.png}
%    \caption{An example where filter response based occlusion
%      detection fails, but depth based occlusion detection can easily
%      handle and keep updating filter without lost track.
%      \label{fig:occlusion}}
%  \end{center}
%\end{figure*}
%

To compensate the weakness of filter response based occlusion handling,
we introduce a depth cue based occlusion detection which is simple
and efficient. Intuitively, all pixels that pass through our
probability based mask generation in Section~\ref{sec:mask} represent
depth values where the target object appears.
We can easily define the amount of occlusion to be allowed by enforcing a threshold
for the visible pixels in $M$
($10\%$ in all our experiments). This depth based occlusion detection
comes without any extra cost since the information is already available from the
masking stage.

Our final occlusion detection combines both the {\em filter response
  based occlusion detection} and {\em depth based occlusion detection}
where an occlusion is declared if both detectors are
triggered, i.e. filter response falls below $65\%$ of moving average
and number of pixels supporting object depth falls as well below
$10\%$ of bounding box regions. If occlusion is detected, the filter
update is stopped and the system switches into full image detection mode (occlusion recovery).
Occlusion recovery model does not make any assumptions on object's reappearance probability 
and it is run as long as the target object is absent in the scene.

%%%%%%%%%%%%%%%%%%%%%%%%%%%%%%%%%%%%%%%%%%%%%%%%%%%%%%%%%%%%%%%%%%%
\section{Experiments}\label{sec:experiments}
In this section, we present an extensive evaluation of the proposed
method. Section~\ref{sec:implementation} provides implementation
details, Section~\ref{sec:dataset} overviews the dataset and the metrics used for the evaluation,
Section~\ref{sec:results} discusses the results and Section~\ref{sec:design} compares different variants of the
proposed method in an ablation study. 

\subsection{Implementation Details}\label{sec:implementation}
To make our results directly comparable to the state-of-the-art, we
selected the same three RGB features in~\cite{csr}
(CSR-DCF): HOG~\cite{Dalal-2005-cvpr}, Color Names~\cite{Weier2009}
and gray level pixel values. We also adopt the same parameter values as in the original
CSR-DCF except
DCF filter update rate ($\psi$) is set to 0.03. Update rates ($\theta$, $\gamma$) 
for probability distributions $P_{fg}$ and $P_{bg}$ are set to $0.95$ and $0.20$
respectively.
% Too much details, ``we use the same'' OR give those different and why
%features. For the parameters, mostly the original CSR-DCF values are
%used while $O$ is empirically set to $0.65$, Gaussian $\mu$ learning
%rate is $0.95$ to make it possible to adapt itself fast, Gaussian 
%$\sigma$ learning rate is $0.20$ to have enough flexibility for
%non-rigid objects and tracking filter learning rate is set to
%$0.03$. For the depth segmentation, $10\%$ of the DCF bounding box is
%chosen as the threshold. 
The parameters were kept fixed in our experiments that were run on
non-optimized Matlab code with Intel Core i7 3.6GHz laptop and Ubuntu
16.04 OS. Our processing speed is calculated according to an average sequence
where the number of occluded frames makes 25\% of all frames.

\subsection{Dataset and Evaluation Metrics}\label{sec:dataset}
Princeton RGBD dataset~\cite{princetonrgbd} consists of
100 sequences from 11 categories and the authors provide ground truth
only for five videos. Methods are evaluated by uploading them to a
specific evaluation server. Results for other methods were taken from
the online leaderboard table at the Princeton website with the exception of
An~\etal~\cite{tlds} who have not registered their method. Therefore,
we took their numbers directly from the respective paper.

Bibi \etal~\cite{Bibi3D} and Hannuna \etal~\cite{DSKCFShape} reported
that $14\%$ of Princeton RGBD dataset videos have synchronization
errors between the RGB and depth frames. In addition, $8\%$ of the
sequences require bounding box re-alignment as pixel correspondences
between RGB and depth frames were erroneous. In their experiments,
Hannuna \etal and Bibi \etal used rectified versions of the dataset
and therefore we found it fair to use their corrected sequences
in our evaluations.

The evaluation uses the widely adopted \textit{Intersection over Union}
(IOU) metric proposed by the authors of the Princeton dataset similar to 
the one used in VOT RGB dataset~\cite{Kristan2016Pami}.

\subsection{Results}\label{sec:results}
\begin{table*}[h]
  \begin{center}
    \caption{Comparison of the best performing methods (online
      evaluation) for Princeton RGBD
      Dataset~\cite{princetonrgbd}.\label{tab:experiments}} 
    \resizebox{1.0\linewidth}{!}{
      \begin{tabular}{lrrrrrrrrrrrrr}
        \toprule
        {\em Alg.} & {Avg Rank} & {\em Human} & {\em Animal} & {\em Rigid} & {\em Large} & {\em Small} & {\em Slow} & {\em Fast} & 
        {\em Occl.} & {\em $\neg$Occl.} & {\em Pass. Motion} & {\em Act. Motion} & \em{FPS}\\
\midrule
3D-T~\cite{Bibi3D}              & 2.81 & \bf{0.81} (1) & 0.64 (4) & 0.73 (5) & \bf{0.80} (1) & 0.71 (3) & 0.75 (5) & \bf{0.75} (1) & \bf{0.73} (1) & 0.78 (5) & 0.79 (3) & 0.73 (2) & N.A\\
RGBDOcc+OF~\cite{princetonrgbd} & 3.27 & 0.74 (4) & 0.63 (5) & \bf{0.78} (1) & 0.78(3) & 0.70 (4) & 0.76 (2) & 0.72 (3) & 0.72 (2) & 0.75 (6) & 0.82 (2) & 0.70 (4) & 0.26\\
OAPF~\cite{oapf}        & 3.45 & 0.64 (6) & \bf{0.85} (1) & 0.77 (3) & 0.73 (5) & 0.73 (2) & \bf{0.85} (1) & 0.68 (6) & 0.64 (6) & \bf{0.85} (1) & 0.78 (4) & 0.71 (3) & 0.9\\
{Our}                     & 3.63 & 0.76 (3) & 0.58 (6) & 0.77 (2) & 0.72 (6) & \bf{\textcolor{forestgreen(web)}{0.73 (1)}} & 0.75 (4) & 0.72 (4) & 0.69 (3) & 0.78 (4) & \bf{\textcolor{forestgreen(web)}{0.82 (1)}} & 0.69 (6) & \bf{\textcolor{forestgreen(web)}{8.3}}\\%13 no-occ, 0.5 occ\\
DLST~\cite{tlds}                & 3.63 & 0.77 (2) & 0.69 (3) & 0.73 (6) & 0.80 (2) & 0.70 (6) & 0.73 (6) & 0.74 (2) & 0.66 (4) & 0.85 (2) & 0.72 (6) & \bf{0.75} (1) & 4.6\\ 		 
DS-KCF-Shape~\cite{DSKCFShape} & 4.18 & 0.71 (5) & 0.71 (2) & 0.74 (4)
& 0.74 (4) & 0.70 (5) & 0.76 (3) & 0.70 (5) & 0.65 (5) & 0.81 (3) & 0.77 (5) & 0.70 (5) & 35.4 \\ %35.4 avg.\\	
CSR-DCF~\cite{csr}             & 10.55 & 0.53 (9) & 0.56 (11) & 0.68 (12) & 0.55 (12) & 0.62 (9) & 0.66 (12) & 0.56 (10) & 0.45 (14) & 0.79 (6) & 0.67 (12) & 0.56 (9) & 13.6\\
        \bottomrule
      \end{tabular}
    }
  \end{center}
\end{table*}

%\vspace{-4\medskipamount}
The results of our and the best performing other trackers for the
Princeton RGBD dataset are given in Table~\ref{tab:experiments}. As it
can be seen below, our method performs on par with the top performing RGBD
trackers (OAPF, RGBDOcc+OF and 3D-T) and is an order of magnitude
faster. Out of the fast trackers (ours, DLST, DS-KCF and
CSR-DCF) ours and DLST are the best with equal average rank, but our
method is faster.
In addition to that, our method wins in two categories:
\textit{Small} and \textit{Passive}. These results indicate that our
depth masked DCF is a suitable tracker
for applications where balance between performance and speed is
important.

The advantages of using the depth channel to complement 2D
information are evident as our DM-DCF outperforms its RGB competitor,
CSR-DCF, in
almost all categories with a clear margin. The only category that
CSR-DCF performs better is the ``no occlusion'' category where benefits
of depth cue are understandably not necessary.
%scenario whereas in the
%presence of occlusion, our approach provides $24\%$ improvement. 
%
Compared to the other DCF based methods DS-KCF-Shape~\cite{DSKCFShape}
and DLST~\cite{tlds}, DM-DCF performs considerably better in
\textit{Occlusion} category. This shows that our occlusion handling
mechanism is more powerful as we use a maximum DCF response score
history in conjunction with foreground segmentation using two separate
probability distributions instead of a single frame response score and
single distribution.  
\subsection{Ablation Study}\label{sec:design}
\begin{figure}[h]
  \begin{center}
    \includegraphics[width=0.8\linewidth]{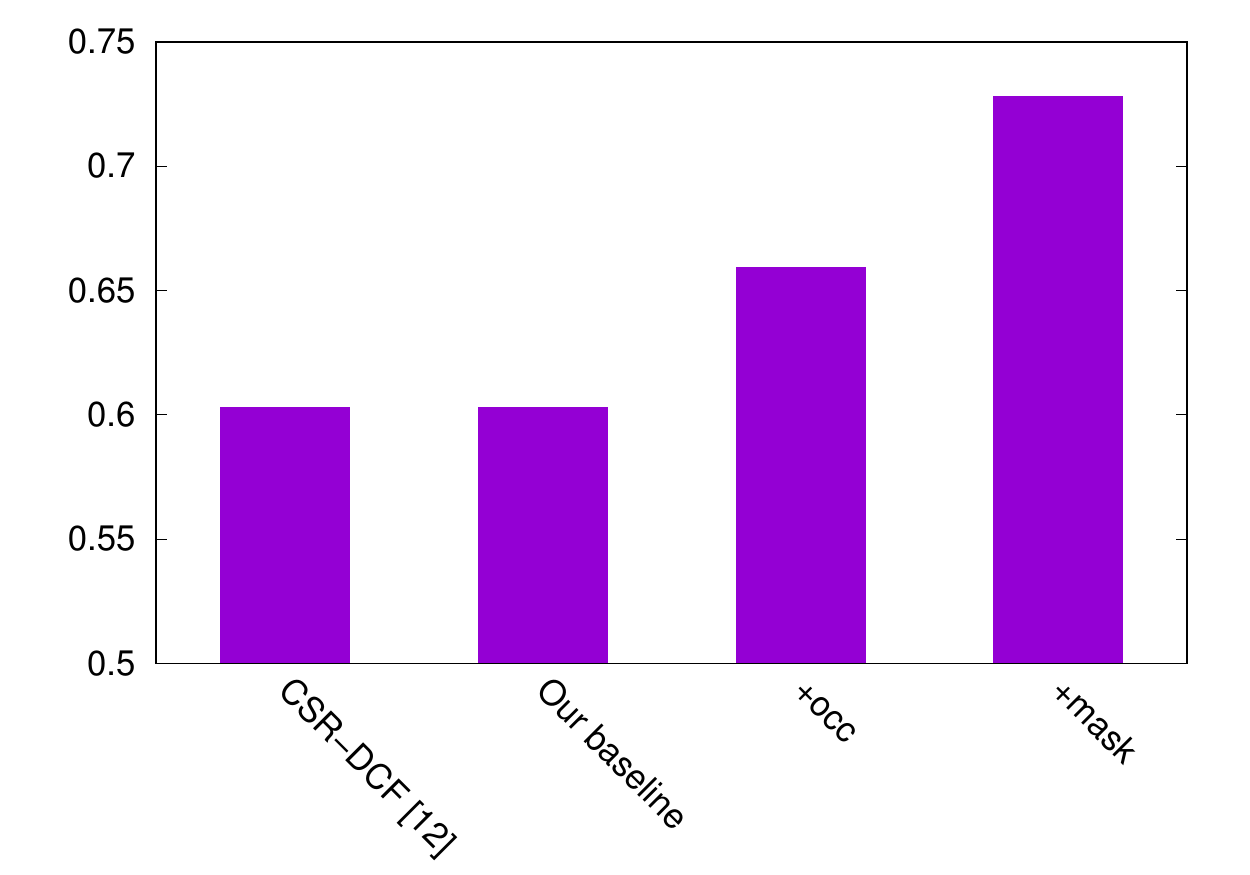}
    \caption{Ablation study of the components of our methods.(Numbers represent
    the average accuracy over 11 categories in the dataset)\label{fig:ablation}}
  \end{center}
\end{figure}
We conducted a set of ablation studies to support our design choices.
Moreover, we also evaluated our algorithm on the original Princeton
Dataset which has considerable amount of registration and
synchronization errors. We report the accuracy for the following
variants:
\begin{compactitem}
\item {\em CSR-DCF} State-of-the-art RGB tracker by Lukezic
  \etal~\cite{csr}
\item {\em DM-DCF--} Our method with the all proposed components switched
  off.
  %\item \textbf{DM-DCF-Incorrect Dataset} is the same algorithm as 
  %DM-DCF, however, it was evaluated on the unrectified Princeton
  %Dataset. - NOT NEEDED
\item {\em +occ} Depth based occlusion handling switched on. %\item
                                %\textbf{DM-DCF-2D Mask} uses the 2D
                                %spatial reliability maps proposed by
                                %Lukezic \etal~\cite{csr} to create a
                                %graph cut based segmentation mask
                                %using color histograms. 
%\item \textbf{DM-DCF-GMM} uses two separate Gaussian Mixture Models
%REF-GMM to model the depth distribution of foreground and background
%instead of single Gaussians. - NOT NEEDED
\item {\em +mask} Depth based masking added i.e. full DM-DCF.
\end{compactitem}
As it can be seen in Fig.~\ref{fig:ablation}, adding depth based masking and occlusion handling
improve the results almost  $\%15$.
\subsection{Summary}
An important finding for the future work is
that in general, different algorithms
favor certain categories and motion types.
As compared the results of all methods in
Table~\ref{tab:experiments}, most of the methods favor rigid motion
over non-rigid motion (rigid vs. animal categories).
This can be explained by the fact that the parameters are
kept constant for all 95 test sequences and the adopted parameters
favors rigid
object tracking. Shape changes and adaptation speeds are different
for non-rigid objects such as animals.

Another similar problem can be seen in occlusion vs. no occlusion.
Again, improvement in the occlusion sequences means slight degradation in 
tracker performance in no occlusion cases. These observations suggest
us to adopt adaptive parameters in our future work so that the tracker
would adjust its parameters on the fly according to the target object.

%%%%%%%%%%%%%%%%%%%%%%%%%%%%%%%%%%%%%%%%%%%%%%%%%%%%%%%%%%%%%%%%%%%
\section{Conclusion}\label{sec:conclusion}
In this paper, we proposed a Depth Masked Discriminative Correlation Filter (DM-DCF)
RGBD tracker that uses the depth cue to detect occlusions (enable switching from
the tracking to the detection mode) and to construct a spatial mask that improves
DCF tracking.
To this end, we are the first to use depth based segmentation
masks inherently in DCF formulation extracting target regions for
filter updates. Comparison and ablation studies on the publicly available Princeton
RGBD Benchmark dataset verified that our trackers is on pair with the
state-of-the-art while providing clearly better frame rate as compared to the
top performers.


%\section*{Acknowledgment}




%\bibliographystyle{IEEEtran}

{\small
\bibliographystyle{ieee}
\bibliography{rgbd_tracking}
}

\end{document}